\documentclass{article}

\usepackage{arxiv}

\usepackage[utf8]{inputenc} % allow utf-8 input
\usepackage[T1]{fontenc}    % use 8-bit T1 fonts
\usepackage{hyperref}       % hyperlinks
\usepackage{url}            % simple URL typesetting
\usepackage{booktabs}       % professional-quality tables
\usepackage{amsfonts}       % blackboard math symbols
\usepackage{nicefrac}       % compact symbols for 1/2, etc.
\usepackage{microtype}      % microtypography
\usepackage{lipsum}
\usepackage{graphicx}
\usepackage{pifont}
\usepackage{algorithmic}
\usepackage{algorithm}
\usepackage{array}
\usepackage{comment}
\usepackage{amssymb}
\usepackage{subcaption}
\usepackage{amsmath}
\usepackage{color}
\graphicspath{ {./images/} }

\title{Federated Unlearning Model Recovery in Data with Skewed Label Distributions}

\author{
 Xinrui~Yu \\
  School of Computer Science and Technology\\ Dalian University of Technology\\ Dalian 116024, China \\
  \texttt{xinruiyu@mail.dlut.edu.cn} \\
   \And
 Wenbin Pei \\
  School of Computer Science and Technology\\ Dalian University of Technology\\ Dalian 116024, China  \\
  \texttt{peiwenbin@dlut.edu.cn} \\
  \And
 Bing~Xue \\
  School of Engineering and Computer Science\\ Victoria University of Wellington\\ PO Box 600, Wellington 6140, New Zealand \\
  \texttt{Bing.Xue@vuw.ac.nz} \\
  \And
  Qiang~Zhang \\
  School of Computer Science and Technology\\ Dalian University of Technology\\ Dalian 116024, China  \\
  \texttt{zhangq@dlut.edu.cn} \\
}

\begin{document}
\maketitle
\begin{abstract}
In federated learning, federated unlearning is a technique that provides clients with a rollback mechanism that allows them to withdraw their data contribution without training from scratch.
However, existing research has not considered scenarios with skewed label distributions. Unfortunately, the unlearning of a client with skewed data usually results in biased models and makes it difficult to deliver high-quality service, complicating the recovery process. 
This paper proposes a recovery method of federated unlearning with skewed label distributions. Specifically, we first adopt a strategy that incorporates oversampling with deep learning to supplement the skewed class data for clients to perform recovery training, therefore enhancing the completeness of their local datasets. Afterward, a density-based denoising method is applied to remove noise from the generated data, further improving the quality of the remaining clients' datasets. Finally, all the remaining clients leverage the enhanced local datasets and engage in iterative training to effectively restore the performance of the unlearning model. Extensive evaluations on commonly used federated learning datasets with varying degrees of skewness show that our method outperforms baseline methods in restoring the performance of the unlearning model, particularly regarding accuracy on the skewed class.
\end{abstract}

% keywords can be removed
%\keywords{First keyword \and Second keyword \and More}

\section{Introduction}
\par {F}{ederated} learning (FL) reduces the risk of data leakage by training models locally on each client’s device and only transmitting model updates to a central server for aggregation \cite{fl,fl2}. In federated learning, clients may withdraw from the training process due to reasons such as protecting their interests, insufficient training resources, or privacy concerns. They may not only cease participating in future training rounds but also request the current federated learning model to ``forget" their data. Therefore, federated unlearning~\cite{ful1,fulsurvey,fulsurvey3} has been proposed to withdraw the data contribution of clients who exit federated learning, without requiring retraining from scratch. This reduces the risk of malicious actors inferring the withdrawn clients' data through reverse engineering. To date, existing studies on federated unlearning \cite{ful2} mainly focus on efficiently removing target clients' contributions from the global model.

\par
\begin{figure}
\centering 
\includegraphics[width=0.5\textwidth]{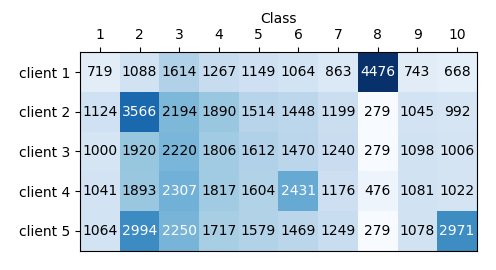}
\caption{Label distribution skew in the federated learning framework. The numbers in the boxes represent the amount of data from each client distributed in 10 classes, and the dark squares represent the skewed class owned by a client.} 

\label{fig1} 
\end{figure}

 Notably, in many real-world applications, one client may have many samples from a certain class, while others have very few. In federated learning, a skewed class refers to a class that is overrepresented in some clients while being underrepresented in others. This causes skewed label distribution among clients \cite{labelskew,labelskew2,fedrs}. For example, in a financial federated learning system, a bank may have substantial transaction data on high-net-worth clients, while small banks have much less. %Similarly, in a federated learning system for hospitals, a certain type of case may be concentrated in one hospital, while other hospitals see fewer of these patients. 
Learning from such data, if a participant withdraws from federated learning and completely removes its data contribution through unlearning algorithms, the model’s performance on that data type can sharply decline, causing severe model performance bias. Figure \ref{fig1} illustrates an example of the skewed label distribution issue among multiple clients. In this example, client 1 has 4476 samples from Class 8, while other clients have fewer. As a result, after the unlearning of Client 1, the performance of the unlearning model on Class 8  will be degraded. However, the existing studies on federated unlearning have not specifically considered the issue of skewed label distribution among clients.

\par Some federated unlearning approaches attempt to restore the quality of the unlearning model by continuing training \cite{fedrs,QuickDrop}. However, in the skewed label distribution, the remaining clients may lack sufficient data for the unlearning class, making it difficult to restore through training effectively. Moreover, most federated learning methods typically tackle the issue of skewed label distribution by adjusting model aggregation or selecting appropriate clients. They aim to increase the contribution of clients with sufficient and high-quality data during the training process. Federated unlearning's privacy requirements prevent organizers from using the data or model of the unlearning client (i.e., the client exiting federated learning), making federated learning and existing recovery methods unsuitable for unlearning model recovery.
This leads to relying on the remaining clients' limited local data to restore the model’s performance bias on the skewed class as much as possible. There are two challenges as follows:

\par \textbf{Challenge 1:} How can we address the performance bias issue caused by skewed label distribution clients? The data from remaining clients is insufficient to maintain model performance. Therefore, under privacy constraints, it remains challenging to address the skewed label distribution issue and effectively restore model quality through training. 
\par \textbf{Challenge 2:} How can we enhance the remaining clients' data by generating high-quality data? The over-sampled data is expected to compensate for the withdrawn data and match the original quality to ensure effective model training.

\par \textbf{Our contributions.} To ensure the quality of the unlearning model under skewed label distribution scenarios, we designed a \textbf{ unlearning model recovery method, named Imba-ULRc.} First, a method combining encoder, decoder, and the synthetic minority over-sampling technique (SMOTE)~\cite{deepsmote} is used to over-sample the skewed class for each remaining client. This allows the data information held by these clients to restore model quality for both the skewed class and overall classes,  addressing the performance bias caused by the skewed label distribution . Then, we further remove noise from the over-sampled data by calculating the density of the data in the feature space, to improve the quality of the over-sampled data. Finally, all the remaining clients use the over-sampled and quality-enhanced local datasets to perform local training on the unlearning model. The local models are then uploaded to the server for aggregation. Through iterative training, the unlearning model ensures reliable high quality across all classes, especially for the skewed class. Specifically, our contributions are as follows:

\begin{itemize}
\item We design a recovery method for the unlearning model. By oversampling the skewed class data, we enrich the data of the remaining clients to make it sufficient for recovery training. This solves the problem of performance bias caused by the skewed label distribution and improves the quality of the unlearning model.
\item 
We design a density-based method to identify noise in the over-sampled data. This further enhances the data quality for the remaining clients and improves the accuracy of the unlearning model.

\end{itemize}

The proposed recovery method is tested on three datasets, each with varying three degrees of skewness, resulting in a total of nine datasets. The received results demonstrated that our method surpasses existing unlearning model recovery algorithms and federated learning algorithms tailored for label-skewed scenarios, achieving higher accuracy for both the skewed class and the overall classes in the recovery model.

\par The remainder of this paper is organized as follows. Section 2 introduces the background of federated unlearning and federated learning with the issue of skewed label distribution. Section 3 introduces our proposed recovery method for the unlearning model under skewed label distribution scenarios. Section 4 introduces experimental designs in detail, and then we verify the performance of the proposed method in Section 5. Section 6 concludes the paper.
\section{Background}
\subsection{Federated Unlearning }
\par The concept of federated unlearning stems from the implementation of ``The right to be forgotten" laws~\cite{gdpr,ccpa}. Its aim is to allow clients who leave the training process to have their contributions removed from the global model. In federated unlearning, a participating client $C_i$ can choose to erase part of its local training data $D_i^{'}$ from its dataset $D_i$. Therefore, it obtains the reduced dataset $D_i^u=D_i\setminus D_i^{'}$, effectively reducing its contribution from the global model~\cite{fedunlearn_define}. Federated unlearning is performed on the global model $\omega_g$ to obtain an unlearning model $\omega_u$, in which the contribution of $D_i^{'}$ should be fully removed. In addition to eliminating the specific data's contribution, the federated unlearning algorithm must preserve the model's overall performance as well. Therefore, the objective of federated unlearning is:

\begin{equation}
    ||\mathbb{T}(\omega_g)-\mathbb{T}(\omega_u)||\leq \varepsilon,
\end{equation}

where $\mathbb{T}$ represents the model distribution, and $\varepsilon$ is a very small positive number. This equation indicates that the unlearning model should have a similar distribution to the federated learning model after unlearning, ensuring no performance deviation. Therefore, federated unlearning algorithms are expected to not only reduce the contribution of the revoked data in the model but also ensure that the model does not suffer from performance bias after unlearning.
\par Studies to date on federated unlearning can be categorized into active unlearning and passive unlearning \cite{ful1}.
\paragraph{Active unlearning} 
In active unlearning, the unlearning client has the right to directly access the data to be forgotten, which can be achieved by executing algorithms on the client itself to withdraw the data contribution. 
In~\cite{UPGA,Wugradient,Forsaken}, gradient correction is used for unlearning. Halimi et al.~\cite{UPGA} propose an unlearning method, where the unlearning client performs gradient ascent on the global model using the data to be erased and controls the degree of unlearning by limiting the distance between the unlearning model and the local models of other clients. Wu et al.~\cite{Wugradient} propose an unlearning method combining elastic weight consolidation and stochastic gradient ascent, where the importance of model parameters is calculated using the Fisher matrix. This method restricts the update of important parameters during gradient ascent. Liu et al.~\cite{Forsaken} propose a method called Forsaken, which deploys a trainable virtual gradient generator for each client and stimulates the model's neurons with virtual gradients during unlearning to eliminate the memory of specific data. 

Differently, some methods~\cite{EXACT,Forget-SVGD} achieve unlearning by retraining. Xiong et al.~\cite{EXACT} propose the Exact-Fun method, where the local model is retrained after the client's data is deleted. If the difference between the newly trained model and the quantized model is large, it will be retrained again. In~\cite{Forget-SVGD}, when a client requests to exit, the unlearning process uses the remaining data to calculate the posterior probability and approximates it through variational inference, finally performing additional rounds of federated learning to achieve model unlearning.

\paragraph{Passive unlearning} The passive unlearning algorithm is executed by the remaining clients and the server, using the server’s cached historical updates of local models or through assistance from the remaining clients to retract the data contributions of the unlearning client. There are approaches~\cite{fedrecovery,Verifi,SIFU} erasing contributions through model fine-tuning and retraining. Zhang et al.~\cite{fedrecovery} introduce removing the influence of an unlearning client by subtracting the sum of weighted gradient residues from the global model and adding Gaussian noise to the unlearning model for additional privacy protection. Gao et al.~\cite{Verifi} propose a method called Verifi, where the server first collects gradient information from all clients, reduces the gradient of the unlearning client, and enhances the gradients of other clients to reduce the influence of the unlearning client. In~\cite{SIFU}, during each round of training, the server calculates the contribution of each client to the global model. When a client requests unlearning, the server retrieves the iteration where the client’s contribution exceeds a predefined threshold from the training history and adds client-specific noise to the intermediate model to achieve unlearning.

\par To date, most unlearning efforts focus on reducing the contribution of the revoked data in the unlearning model, with only a few considerations on recovery training after unlearning~\cite{FUDP,QuickDrop}. However, under the scenario with the issue of label distribution skew, such recovery methods struggle to restore the unlearning model to a distribution that is similar to that of the global model, leading to performance degradation in the unlearning model.
\subsection{Federated Learning with Skewed Label Distribution 
}
\par Skewed label distributions usually occur in many real-world scenarios. In federated learning, since data is confined to local clients, the difficulty of handling data imbalance in the federated setting increases~\cite{flnon-iidsurvey1,flnon-iidsurvey2,flnon-iidsurvey3,flnon-iidsurvey4}. Improving federated aggregation could alleviate model performance bias \cite{scaffold,moon}. 
In \cite{scaffold,moon}, local model drift is corrected by making the update direction of each client closer to the global model update direction. 
Additionally, other works address this imbalance by selecting appropriate clients. In \cite{DOCS,Zhang}, the imbalance issue in federated systems is addressed by adaptively selecting participants or choosing suitable participants based on data similarity across clients. Directly improving classifiers is another approach to preventing performance bias. Liu et al.~\cite{applied1imbalance} propose an attention mechanism-based aggregation method to improve global model performance and dynamically select clients to ensure that each client’s characteristics are reflected in the global model. In~\cite{zhaoshare,shareItahara,shareTian}, a small-scale globally shared dataset is used, allowing all client devices to use this data for training.
%\textcolor{cyan}{Zhao et al.~\cite{zhaoshare}, Itahara et al.~\cite{shareItahara}, and Tian et al.~\cite{shareTian} adopted a small-scale globally shared dataset, allowing all client devices to use this data for training.} 
Li et al.~\cite{fedrs} propose a method, where the softmax layer in the model is replaced with a constrained softmax, to reduce the negative impact on class weights caused by missing samples locally. Ran et al.~\cite{DMFL} introduce a dynamic margin loss, and set a larger margin for minority class samples, thus improving the convergence speed and overall test accuracy in federated learning on skewed data.
\par These methods can reduce performance bias caused by skewed label. However, addressing the performance bias in federated unlearning still remains challenging, because clients with a large amount of skewed class data exit the training, causing the remaining clients to lack data for the skewed class. This situation renders many federated learning methods for handling label skew unsuitable for recovery. Therefore, it motivates us to design recovery methods specifically for federated unlearning.

\section{The Proposed method: Imba-ULRc}
\begin{figure*}[!t] 
    \centering
    \includegraphics[width=\textwidth]{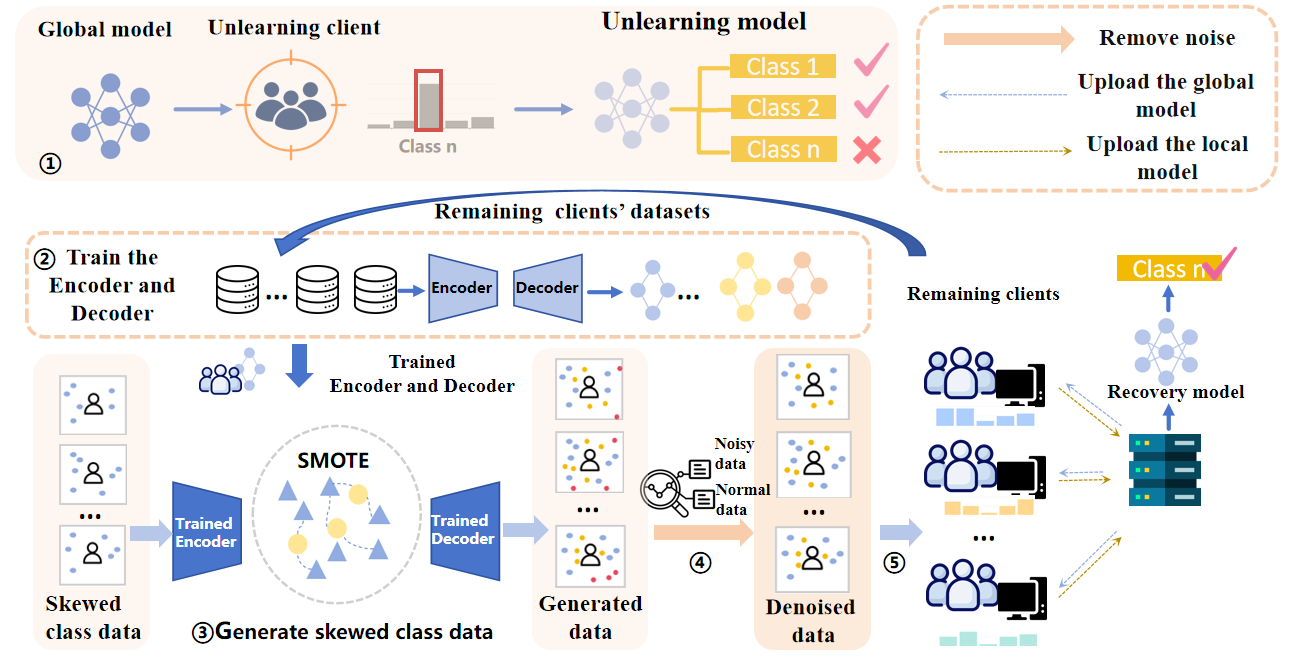} 
  
    \caption{The overview of the proposed method. \ding{192} indicates the unlearning model's performance bias caused by the skewed class's presence in the skewed label distribution.
\ding{193} and \ding{194} represent our proposed method for Addressing the Performance Bias. First, we train a data generation model composed of an encoder and decoder and then use SMOTE to oversample the skewed class of the remaining clients.
\ding{195} and \ding{196} represent our proposed Data Quality Enhancement and Recovery Training method. First, we remove noise from the generated data of each client, and then, through iterative training with the server, we finally obtain a high-quality recovery model.} 
    \label{fig:Framework}
\end{figure*}
\subsection{The Overview of the Proposed Method}

\par To tackle the challenges of quality deterioration and recovery difficulties in an unlearning model with skewed label distribution contexts, we propose a recovery method of the unlearning model, called Imba-ULRc. The overall design of Imba-ULRc is indicated in Figure \ref{fig:Framework}, where Imba-ULRc has three important phases, introduced below.

\textbf{The first is to address the performance bias} by over-sampling the skewed class data of remaining clients, as shown in Steps \textcircled{2} and \textcircled{3} in Figure \ref{fig:Framework}. Each remaining client uses its complete local data to train an encoder that reduces data dimensions. SMOTE is then used to over-sample the data in the reduced feature space. Afterward, the decoder is trained to decode the data into the original data representation.

\textbf{The second is to enhance the data quality} by removing noise from the generated data, as shown in Step \textcircled{4} of Figure \ref{fig:Framework}. For each generated data point, the density within the entire skewed class is calculated. Next, {$k$} nearest neighbors of each generated data point within the skewed class are identified in the original feature space, and their densities are calculated to determine the average density of these neighbors for each generated data point. The density factor of a generated data point is calculated by comparing its density with the average density of its neighbors. 
Data points with a density factor exceeding a predefined threshold are removed, ensuring both data quality and quantity.

\textbf{The third is to use the enhanced local datasets by the remaining clients for recovery training}, as shown in Step \textcircled{5} of Figure \ref{fig:Framework}.  The server coordinates the remaining clients to engage in recovery training. The unlearning model is assigned to each client, where it is locally trained on the enhanced datasets. After completing local training, each client uploads its local recovery model to the server. The server consolidates all these local models and continues training them iteratively until the global model converges. This process is expected to develop an enhanced recovery model, effectively facilitating recovery of the unlearning model in scenarios with skewed label distributions.

\begin{comment}

\end{comment}
\subsection{Addressing the Performance Bias}

\par In federated learning, the issue of skewed label distribution is well-studied \cite{labelskew2}. Building on this, we propose the concept of data skew specifically within the context of federated unlearning.
\\

 \textbf{Definition 1 (Skewed data distribution in federated unlearning).} Each client has a local data distribution $P(x,y)$, where the data distribution of the unlearning client is denoted as $P_u(x,y)=P_u(x|y)P_u(y)$. This distribution is different from $P_i(x,y)=P_i(x|y)P_i(y)$ for remaining clients $i\neq u$. Let the majority class be $\mathcal{J}$ and the skewed class owned by the unlearning client be $\mathcal{C}$, the set of all classes is denoted as $\mathcal{K}$. Note that $\mathcal{J}\cup\mathcal{C}=\mathcal{K}$ and $\mathcal{J}\cap\mathcal{C}=\varnothing$. The amount of data that the unlearning client has in the skewed class is $n_\mathcal{C}^u$ while remaining clients have $n_\mathcal{C}^i$ in the skewed class and $n_\mathcal{J}^i$ in the majority class, with $n_\mathcal{C}^u$ $\gg$ $n_\mathcal{J}^i$$\gg$$n_\mathcal{C}^i$.
\\

\par 
The proposed Imba-ULRc method aims to address performance bias and thereby improve the accuracy of the unlearning model on the skewed class $\mathcal{C} $ after recovery while maintaining the accuracy of the global recovery model. To achieve this goal, we first oversample the skewed class on the remaining clients, ensuring sufficient data information to restore the model, as indicated in Steps \textcircled{2} and \textcircled{3} of Figure \ref{fig:Framework}. % ensuring that their data information is sufficient to restore the model.

\par \textbf{First}, in the remaining clients, local resources are used to train an encoder-decoder model with the capability of generating data. In more detail, for each remaining client $i$, represented as $C^r_i$, the complete data $D_i^r$ is utilized to train the encoder $\mathbb{Z}_{en}$. The $h$-th data point from client $i$ is denoted as $d^h_i \in D_i^r$. A $d^h_i$ is passed through the encoder to generate a low-dimensional representation, denoted as $x^h_i=\mathbb{Z}_{en}(d^h_i)$. The obtained low-dimensional representation is fed into the decoder $\mathbb{Z}_{de}$ for data reconstruction, denoted as $y^h_i=\mathbb{Z}_{de}(x^h_i)$. Each client then computes the reconstruction loss by:
\begin{equation}
    \mathbb{L}_1=\frac{1}{n^i}\sum_{h=1}^{n^i}(y^h_i-d^h_i)^{2},\label{restruct_loss}
\end{equation}
where $n^i$ is the total amount of data from the client $i$. This enables the decoder to reconstruct the reduced-dimensional features back into the feature space of the original data.

\par Moreover, each remaining client rearranges the decoder's inputs with reduced dimensionality to introduce variations in the feature space, allowing the decoder to reconstruct the altered data with reduced dimensionality. Specifically, each client $i$ randomly selects a class $c$ and inputs the data of class $c$ sequentially into the encoder during training. We then obtain the sequential encoder outputs $x_{i,c}^1,x_{i,c}^2,...,x_{i,c}^p=\mathbb{Z}_{\mathrm{en}}(d_{i,c}^1,d_{i,c}^2,...,d_{i,c}^p)$, where $d_{i,c}^p$ is the $p-th$ data of class $c$ for client $i$, and $x_{i,c}^p$ is its encoded result. During decoding, the reduced representations from the encoder are input into the decoder in reverse order. In other words, the reduced representations generated by the encoder in sequence, i.e., $x_{i,c}^1, x_{i,c}^2, ..., x_{i,c}^p$, are fed into the decoder in the opposite order. It becomes $x_{i,c}^p, ..., x_{i,c}^2, x_{i,c}^1$, producing $y_{i,c}^p, ..., y_{i,c}^2, y_{i,c}^1 = \mathbb{Z}_{de}(x_{i,c}^p, ..., x_{i,c}^2, x_{i,c}^1)$. Here, $y_{i,c}^p$ is the result of the decoder reconstructing $x_{i,c}^p$. By comparing these outputs with the original data sequence, the loss function $\mathbb{L}_2$ is calculated.:
\begin{equation}
    \mathbb{L}_2=\frac{1}{p}[(y^p_{i,c}-d^1_{i,c})^{2}+(y^{(p-1)}_{i,c}-d^2_{i,c})^{2}+...+(y^1_i-d^p_{i,c})^{2}]. \label{reLoss2}
\end{equation}
Therefore, the target loss function for each client during training is $\mathbb{L}=\mathbb{L}_1+\mathbb{L}_2$. It enables the local data generation model to produce high-quality data and generate data even when the reduced low-dimensional feature space differs.

\par \textbf{Second}, each client leverages its locally trained encoder-decoder model along with SMOTE to over-sample the skewed class in the reduced feature space. For a sample of the skewed class, the client locates the nearest neighbors in the reduced-dimensional feature space. Then, in the feature space, it interpolates between the sample and one of its neighbors, creating new samples. Specifically, a new sample in the reduced feature space is constructed by the equation as follows:
\begin{equation}
   x^l_{i,g} = x^l_{i,\mathcal{C}} + \mathnormal{Rand}(0,1)*(x^l_{i,N}-x^l_{i,\mathcal{C}}),
\end{equation}
where $x^l_{i,\mathcal{C}}$ represents the $l$-th data point of the skewed class $\mathcal{C}$. $x^l_{i, N}$ is the neighboring node of $x^l_{i,\mathcal{C}}$ in the reduced low-dimensional feature space, $x^l_{i,g}$ is the reduced-dimensional representation of the newly over-sampled data, and $Rand(0,1)$ is a random number distributed between 0 and 1. Then, $x^l_{i,g}$ is input into the decoder, $d^l_{i,g} = \mathbb{Z}_{de}(x^l_{i,g})$, to obtain the new sample $d^l_{i,g}$. Thus, this method allows the remaining clients to generate data for the skewed class. 
\subsection{Data Quality Enhancement}

\par To remove noise in the generated data, we apply a density-based method.
In the last phase, SMOTE is used in a low-dimensional space to synthesize data by introducing random variations between generated samples and their neighbors. These variations are then processed through the decoder to create data with the same dimensions as the original dataset. However, a challenge arises because the variations in the low-dimensional representations introduced by $Rand(0,1)$ in SMOTE are random, and the decoder may not always decode these low-dimensional representations into high-quality original data. This could result in noise in the generated data. 

\par To mitigate this, we first calculate the density in the original feature space of the data. For client $i$'s each generated data point $d_{i,g}^j$, we identify its {$k$} nearest neighbors among the same-class data (denoted as $\mathcal{N}_q(d_{i,g}^j)$,  where $q=1,...,k$), and calculate the Euclidean distance between data point $d_{i,g}^j$ and each $\mathcal{N}_q(d_{i,g}^j)$, denoted as $E(d_{i,g}^j, \mathcal{N}_q(d_{i,g}^j))$. Note that the same-class data encompasses both the original and generated data from the skewed class. After obtaining the distances to the neighbors, we calculate the density $\phi_{i,g}^j$ of the $j$-th generated data for client $i$ by the following formula:
\begin{equation}
    \phi_{i,g}^j=(1+\frac{\sum_{\mathit{q}=1}^k E(d_{i,g}^j, \mathcal{N}_q(d_{i,g}^j))}{K+1})^{-1}.
\end{equation}

Thus, each generated data point's density within the same-class data is calculated based on the distance to its surrounding neighbors. Note that the smaller the distance to its neighbors, the higher the density.

\begin{figure}
  \begin{minipage}[t]{0.5\linewidth}
    \centering
    \begin{subfigure}[t]{\linewidth}
      \centering
      \includegraphics[scale=0.31]{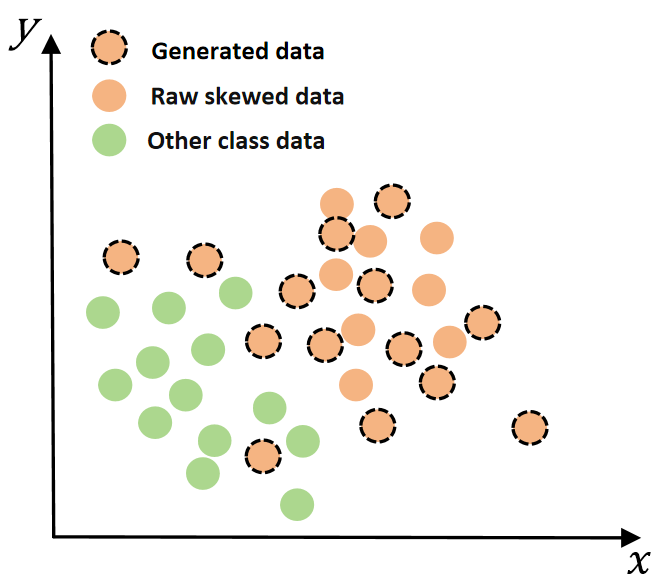}
      \caption{ Skewed class data with noise.}
      \label{desitya}
    \end{subfigure}
  \end{minipage}%
  \begin{minipage}[t]{0.5\linewidth}
    \centering
    \begin{subfigure}[t]{\linewidth}
      \centering
      \includegraphics[scale=0.15]{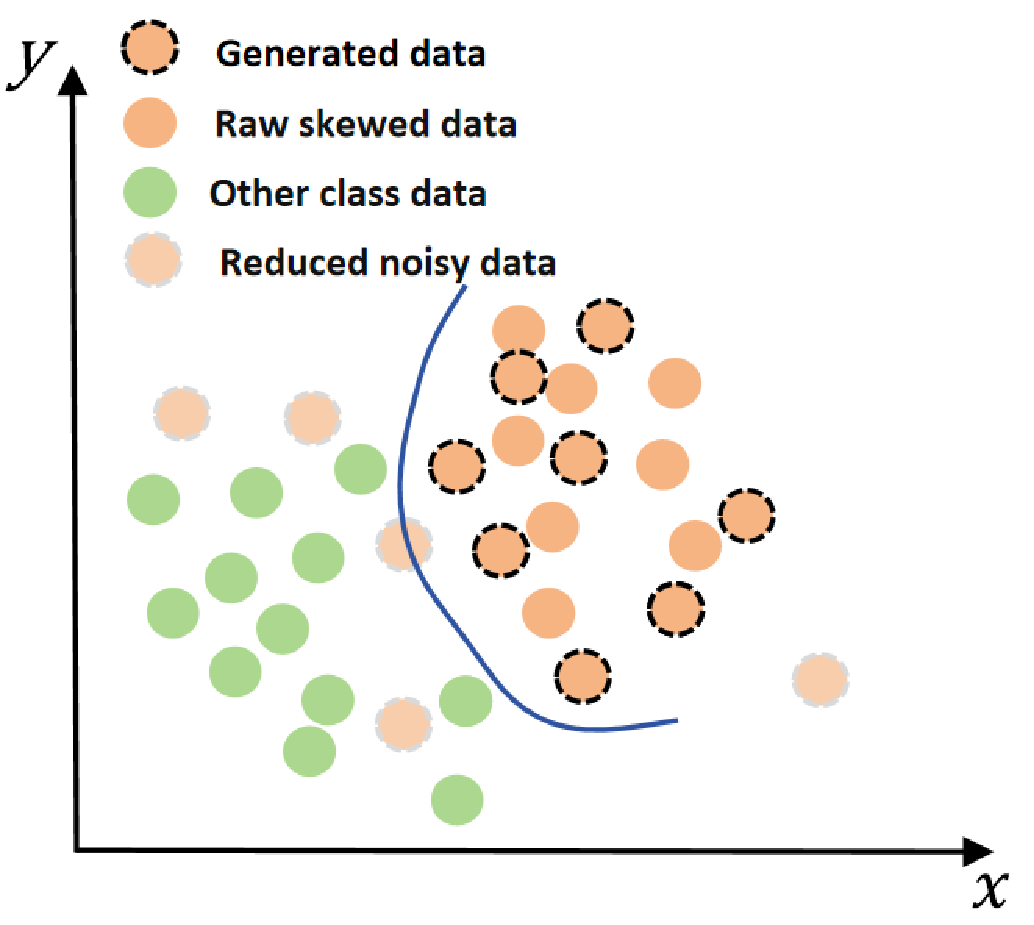}
      \caption{ Denoised skewed class data. }
      \label{desityb}
    \end{subfigure}
  \end{minipage}
  \caption{The density-based method for removing noise from generated data. The orange dots represent skewed class, the green dots represent other majority class, and the dashed lines indicate the generated skewed class data.}
\end{figure}

\par Usually, data points sharing the same label are positioned closer together in the feature space, as indicated in Figure \ref{desitya}. However, some data from the oversampling phase may deviate from the class center, causing blurred boundaries between classes. Such noisy data exhibit lower density in the feature space relative to nearby same-class data. Thus, by evaluating and comparing the density of the generated data with that of its neighboring data, this noise can be detected and removed, making the class decision boundaries clearer, as shown in Figure \ref{desityb}. The density factor of the generated data is calculated as:

\begin{equation}
    \Psi_{i,g}^j =\frac{\sum_{q=1}^k(1+\frac{\sum_{\mathit{q^{'}}=1}^k E(\mathcal{N}_{q}, \mathcal{N}_{q')} }{k+1})^{-1}}{k\phi_{i,g}^j},\label{density_ratio}
\end{equation}
where $\mathcal{N}_{q}$=$\mathcal{N}_q(d_{i,g}^j)$ denotes the neighbors of $d_{i,g}^j$, $\mathcal{N}_{q'}$ is the neighbors of data $\mathcal{N}_{q}$. The numerator represents the sum of the densities of data $j$'s neighbors. The density factor for data point $j$ is obtained by dividing the average density of data point $j$'s neighbors by the density of data point $j$.

\par  We then use the obtained density factor to determine whether a generated sample is removed from the generated data or not. When comparing a point's local density to that of other points in its neighborhood, most points have densities similar to their neighbors~\cite{lof,chandola2009anomaly,papadimitriou2003loci}. This leads to density factors predominantly concentrated in the left region of the distribution, with values clustering around 1, while a few points exhibit densities extending further to the right.
This indicates that the distribution of density factors exhibits a right-skewed pattern. For skewed data distributions, the mean of the obtained density factors is sensitive to extreme values. In contrast, the median value more accurately represents the data's central tendency and is less affected by outliers. Therefore, we choose the median of the obtained density factors as the threshold for noise removal.

In Imba-ULRc, we use SMOTE to generate twice as many samples as needed for supplementing a amount of data for the skewed class. Afterward, a generated sample will be removed if its density factor is greater than the median of all the other samples' density factors. This removal process continues until the data reaches balance. Finally, each client obtains a denoised dataset $D_{dn}^i$ through this process.

\begin{algorithm}[H]
       
	 \textbf{\textit{Remaining client update}}
        \renewcommand{\algorithmicrequire}{\textbf{Input:}}
	\renewcommand{\algorithmicensure}{\textbf{Output:}}
	\caption{Recovery of the unlearning model in skewed label data distribution scenarios}
	\label{alg:1}
	\begin{algorithmic}[1]
		\REQUIRE Skewed dataset $D^r_i$ of Client $i$, the unlearning model $\omega_u$.
		\ENSURE   The local recovery mode $\omega_i^{t}$ of the remaining client.

	\STATE Train the encoder $x^h_i=\mathbb{Z}_{en}(d^h_i)$ and the decoder $y^h_i=\mathbb{Z}_{de}(x^h_i)$, and compute the reconstruction loss according to equation \eqref{restruct_loss}.

        \STATE Change the input order of the decoder and compute the reverse reconstruction loss according to equation \eqref{reLoss2}.

        \STATE Use SMOTE and the trained models $\mathbb{Z}_{en}$ and $\mathbb{Z}_{de}$ to generate data for the skewed class.
	
        \STATE Calculate the density factor for the generated data according to equation \eqref{density_ratio}.
        
        \STATE Remove the generated data with high-density factors to gain a balanced dataset $D_{dn}^i$.
        \FOR {local update rounds 1 to T}
        
            \STATE Train the local model using the enhanced dataset $D^i_{dn}$ 
            \\$\omega_i^{t}=\omega_u-\eta\nabla\mathcal{L}(\omega_u,D^i_{dn})$
        \ENDFOR
        \STATE Send $\omega_i^{t}$ to the server.
       
        \STATE Continue local training upon receiving the global recovery model $\Omega^t$.
	\end{algorithmic}  

      \textbf{\textit{Server update}}
       \renewcommand{\algorithmicrequire}{\textbf{Input:}}
	\renewcommand{\algorithmicensure}{\textbf{Output:}}
 	\begin{algorithmic}[1]
       
	\REQUIRE The local recovery model $\omega_i^{t}$ sent by the remaining clients.
       
		\ENSURE Global recovery model $\Omega$.
            \FOR{global update rounds 1 to $\tau$}
            
	    \STATE Aggregate the received local recovery models $\omega_i^{t}$ according to equation \eqref{aggregation} to obtain the global recovery model for the current round $\Omega^t$.
     
            \STATE send $\Omega^t$ to remaining clients.
            \ENDFOR
       
            \STATE Obtain the global recovery model $\Omega$.
	\end{algorithmic} 
 \label{a1}
\end{algorithm}

\subsection{Recovery Training}
\par
\begin{figure}
\centering 
\includegraphics[width=0.6\textwidth]{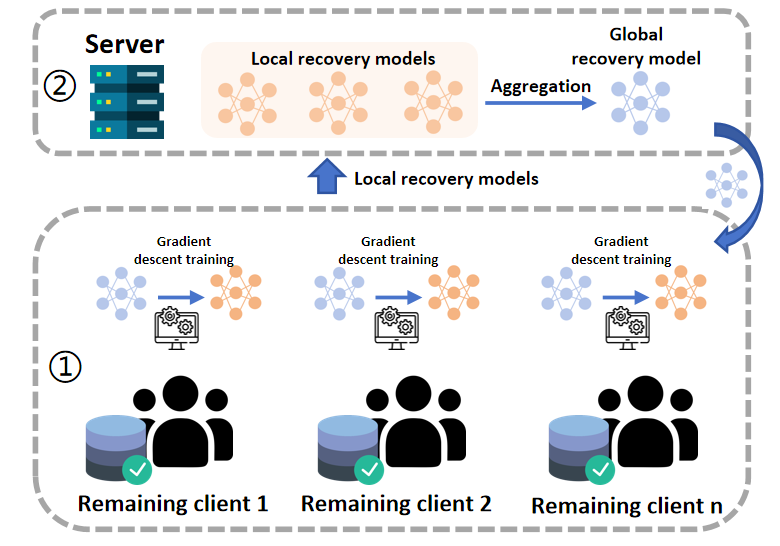}
\caption{Details of the recovery training.}

\label{recovery training} 
\end{figure}

 The remaining clients obtain a complete local dataset that includes additional data for the skewed class, with noise in the newly generated data removed. These clients are then responsible for integrating the knowledge of the enhanced skewed class into the unlearning model. This ensures that the model retains better performance, preserving its overall usability even if clients leave. The specific details of recovery training are shown in Figure~\ref{recovery training}.

\par The server coordinates all remaining clients to conduct recovery training on the unlearning model using enhanced local datasets. Server $S$ distributes the unlearning model $\omega_u$ to the remaining clients, and client $i$ utilizes its local computational resources to perform gradient descent training on its dataset $D^i_{dn}$, as shown in Figure~\ref{recovery training} $\textcircled{1}$. Therefore, the local recovery model $\omega_i^{t}$ is obtained:
\begin{equation}
\omega_i^{t}=\omega_u-\eta\nabla\mathcal{L}(\omega_u,D^i_{dn}),
\end{equation}
where $\eta$ represents the learning rate. In Figure~\ref{recovery training} $\textcircled{2}$, once all remaining clients have finished their training, the server aggregates the received models to generate the global model $\Omega^{t}$ for the $t$-th round. This is defined as follows:
\begin{equation}
\Omega^{t}=\sum_{i=1}^M\frac{N^{dn}_i}{\sum_{j=1}^M N^{dn}_j}\omega_i^t,\label{aggregation}
\end{equation}
where $N^{dn}_i$ denote the total amount of noised dataset $D^i_{dn}$  for client $i$, allowing to acquire the global aggregated model $\Omega^{t}$, $M$ represents the number of remaining clients. The server then sends the aggregated model back to all remaining clients for further local training. After completing this round of training, the server performs aggregation again. This process is repeated for $\tau$ rounds, ultimately producing the restored model $\Omega$.

\subsection{Training Process of the  Recovery Method}

\par Algorithm~\ref{a1} shows the detailed training process of the proposed recovery method. %During training, we combine the methods of addressing performance bias with noise removal. 
After completing the encoder-decoder training, we generate a certain amount of data, remove noise from it, and then use the refined data for local training. The server is responsible for distributing models and aggregating during training, ultimately obtaining a global recovery model on the server side.

\section{Experimental details}

\par 
In this section, we introduce datasets and models utilized in the experiments along with the hardware configuration. Then, we introduce the experiment setup and data allocation for the federated unlearning and remaining clients, and the experimental pipeline. The baseline recovery methods are also detailed.

\subsection{Datasets and Models}

\par In this study, we select three commonly used image datasets in federated learning, each configured with three different degrees of skewness, resulting in a total of nine datasets to evaluate the effectiveness of the proposed method. The MNIST \cite{MNIST} dataset, a prevalent collection of handwritten digit images, comprises 60,000 training images and 10,000 test images. FMNIST \cite{fmnist} includes 70,000 grayscale images of fashion items across 10 classes. %, acts as a challenging successor to the traditional MNIST dataset. 
The USPS \cite{USPS} dataset includes  7291 training grayscale images and 2007 test images of digitized postal codes from the U.S. Postal Service. These image datasets are widely used in federated learning. 

\par To test the algorithm's performance under varying levels of skewness, for each original dataset, we randomly select a class as the skewed class for the unlearning client and form three skewed data scenarios by allocating varying proportions of data from the skewed class to the unlearning client. In the experiments, we set $\alpha=\{0.80,0.85,0.90\}$ to indicate the unlearning client holding 80\%, 85\%, and 90\% of the total data in the skewed class, respectively. After that, we distribute the remaining data of the skew class, along with the data from other classes, evenly among the other clients. This ensures that the total amount of data for each client remains consistent.

\par  For MNIST and FMNIST, We utilizes a multi-layer perceptron (MLP) model, comprising an input layer, a hidden layer activated by ReLU, enhanced with dropout regularization, and a final output layer. In the case of USPS, we implement a convolutional neural network (CNN) model. It includes two convolutional layers, each followed by max pooling and ReLU activation, with additional dropout regularization after the second convolution. Finally, two fully connected layers are used for classification.

\par We configured the training with a batch size of 64 and a learning rate of 0.01, utilizing the stochastic gradient descent optimizer with a Momentum of 0.5. For testing, the batch size is set to 128, and cross-entropy is employed as the loss function. The experiment was performed in a computing environment featuring an NVIDIA Tesla V100 GPU. To simulate federated training on a single machine, we perform a series of client-side local training sequentially.

\subsection{Federated Learning Setting and Pipeline}

\textbf{FL-Setting:} We configure five federated learning clients and treat the first client (Client 1) as the unlearning client, while the other clients participate as the remaining clients in the recovery training. In this setup, we introduce a skew in label distribution among the clients. All remaining clients train \textcolor{black}{all datasets} locally for $T=2$ rounds, meaning that they upload their models to the server after two rounds of local training. The number of global training rounds is $\tau=10$ for all datasets and methods, indicating that the server performs a total of ten aggregations of the local models. 

\par \textbf{Pipeline:} We briefly introduce the entire process of the experiment, from federated training to unlearning and then to recovery.

\begin{itemize}	
	\item \textbf{Training before recovery:}  Prior to recovery training, it is necessary to engage in federated learning and unlearning training to acquire the foundational model for subsequent training. All client nodes undertake federated learning, conducting two rounds of local training per session. Once the global model converges, federated training is halted by the server. Given that the selection of an unlearning algorithm does not impact the outcome of the experiment, the \textbf{U}nlearning with \textbf{P}rojected \textbf{G}radient descent \textbf{A}lgorithm (\textbf{UPGA})~\cite{UPGA} is employed as the representative for conducting unlearning trials. The unlearning client apply gradient ascent using all available data on their local compute resources to the received global model, thus deriving the unlearning model.
	
  \item \textbf{Recovering the unlearning model:} The remaining clients proceed with recovery training, employing the proposed Algorithm 1 to refine local data and remove noise. Subsequently, they perform local recovery training on the unlearning model. Each local training phase includes two rounds, continuing until the global recovery model converges and the training is discontinued.
  
\end{itemize}

\subsection{Baseline Methods}
\par We compare our proposed unlearning model recovery method with two unlearning fine-tuning methods in federated unlearning, as well as two federated learning methods designed for imbalanced data scenarios. The four baseline methods are detailed as follows:
%\par \textbf{Baselines:}
%\par 
\begin{itemize}
\item
   
    The \textbf{F}ederated \textbf{U}nlearning via class-\textbf{D}iscriminative \textbf{P}runing~(\textbf{FUDP}) method~\cite{FUDP}: 
This method quantifies the class discriminability of channels using the Term Frequency-Inverse Document Frequency (TF-IDF), prunes channels to unlearn specific classes and then fine-tunes the model through retraining. We compare our method with FUDP's retraining recovery method.
\item
   
    The efficient federated unlearning by integrated dataset distillation~\cite{QuickDrop} (named as \textbf{QUICKDROP}) method: %Dhasade et al. developed a technique to 
    This method hastens the processes of unlearning and recovery by distilling large datasets on each client into smaller ones. This compression ensures that the essential features of the original data are preserved in the smaller versions, which are then utilized for both unlearning and recovery. We compare our proposed method with QUICKDROP's recovery method. 
    \item
    
   The \textbf{F}ederated \textbf{L}earning with \textbf{R}estricted
\textbf{S}oftmax~(\textbf{FLRS}) method~\cite{fedrs}:
   This method mitigates inaccuracies in softmax layer updates caused due to the skewed class by limiting updates during local training. Note that the method was originally designed for federated learning. To compare with this method, we build the unlearning model by incorporating their proposed constrained softmax classification model during recovery training.
\item 

The Model-contrastive federated learning~(named as \textbf{MOON}) method~\cite{moon}: 
In MOON, local updates are corrected by maximizing the consistency between the representations learned by the current local and global models through model-level contrastive learning. This method effectively enhances the performance of federated learning for imbalanced data classification tasks. To build the MOON-based unlearning recovery model, we modify the local recovery training loss function for comparison.

\end{itemize}

\section{Results and analysis}
This section consists of three major parts:  
(1)~comparisons between the proposed method and the baseline methods; (2) ablation studies on the proposed method;
(3) parameter sensitivity analysis.
\subsection{Comparisons with Baseline Methods}

\begin{table*}
    \captionsetup{format=plain, textformat=period, labelsep=newline, justification=centering}

    \caption{The Accuracy of the Skewed Class and \textbf{Balanced Accuracy} of the Global Model Under Different Levels of Skewness}
    \centering
    
    \scalebox{0.8}{
    \begin{tabular}{ccccccccc}
    \toprule
    Dataset & Recovery approach & \multicolumn{2}{c}{$\alpha=0.8$} & \multicolumn{2}{c}{$\alpha=0.85$} & \multicolumn{2}{c}{$\alpha=0.9$} \\ 
    \cmidrule(r){3-4} \cmidrule(r){5-6} \cmidrule(r){7-8}
    & & Skewed class  & Global model  & Skewed class  & Global model  & Skewed class  & Global model  \\ 
    \midrule
MNIST & FUDP & $88.85 _{\pm 0.45}$ &  $95.52 _{\pm 0.02}$ & $86.38_{\pm 0.29}$ & $95.47 _{\pm 0.26}$ & $81.61 _{\pm 0.66}$ & $94.72 _{\pm 0.31}$ \\
& QUICKDROP & $87.09 _{\pm 1.00}$ & $91.12 _{\pm 0.11}$ & $86.80 _{\pm 1.03}$ & $91.34 _{\pm 0.08}$ & $86.71 _{\pm 1.13}$ & $91.33 _{\pm 0.06}$ \\
& FLRS & $91.28 _{\pm 0.23}$ & $93.14 _{\pm 1.78}$ & $88.35 _{\pm 0.48}$ & $93.09 _{\pm 0.73}$ & $87.93 _{\pm 1.36}$ & $94.61 _{\pm1.94}$ \\
& MOON & $88.78 _{\pm 0.20}$ & $95.62 _{\pm 0.04}$ & $87.39 _{\pm 0.15}$ & $95.69 _{\pm 0.05}$ & $81.61 _{\pm 0.64}$ & $95.09 _{\pm0.06}$ \\

& \textbf{Imba-ULRc } & \textbf{$\textbf{93.38} _{\pm 0.33}$} & \textbf{$\textbf{95.82} _{\pm 0.02}$} & \textbf{$\textbf{93.06} _{\pm 0.34}$} & \textbf{$\textbf{95.94} _{\pm 0.21}$} & \textbf{$\textbf{92.41} _{\pm 0.68}$} & \textbf{$\textbf{95.91} _{\pm 0.03}$} \\
\midrule

FMNIST & FUDP & $80.18 _{\pm 0.14}$ & $85.29 _{\pm 0.11}$ & $74.42 _{\pm 0.21}$& $84.90 _{\pm 0.14}$ & $71.23 _{\pm 0.12}$ & $84.60 _{\pm 0.18}$\\
& QUICKDROP & $80.53 _{\pm 0.84}$ & $80.02 _{\pm 0.20}$ & $80.57 _{\pm 2.35}$& $79.98 _{\pm 0.19}$ & $77.70 _{\pm 2.08}$ & $79.62 _{\pm 0.47}$\\
& FLRS & $87.70 _{\pm1.05}$ & $85.58 _{\pm 0.28}$ & $87.43 _{\pm 0.67}$& $85.86 _{\pm 0.11}$ & $80.87 _{\pm 0.47}$ & ${85.46} _{\pm 0.11}$\\
& MOON & $77.60 _{\pm 0.56}$ & $84.93 _{\pm 0.04}$ & $76.67 _{\pm 1.33}$ & $85.23 _{\pm 0.09}$ & $65.77 _{\pm 0.64}$ & $84.12 _{\pm0.17}$ \\
&  \textbf{Imba-ULRc }  & $\textbf{89.30} _{\pm 0.36}$ & $\textbf{85.75} _{\pm 0.14}$ & $\textbf{89.77} _{\pm 0.40}$& $\textbf{86.12} _{\pm 0.08}$ & $\textbf{89.07} _{\pm 0.23}$ & $\textbf{86.06} _{\pm 0.04}$\\
\midrule

USPS & FUDP & $73.73 _{\pm 0.36}$ & $84.99 _{\pm 0.15}$ & $66.45 _{\pm 0.08}$& $84.14 _{\pm 0.72}$ & $58.17 _{\pm 0.29}$ & $81.03 _{\pm 0.08}$\\
& QUICKDROP & $82.84 _{\pm 0.70}$ & $63.15 _{\pm 0.55}$ & $81.23 _{\pm 0.20}$& $64.85 _{\pm 0.51}$ & $80.86 _{\pm 0.34}$ & $61.38 _{\pm 0.06}$\\
& FLRS & $85.57 _{\pm0.23}$ & $\textbf{85.50} _{\pm 0.09}$ & $78.22 _{\pm 0.01}$& $84.60 _{\pm 0.35}$ & $70.18 _{\pm 0.20}$ & $\textbf{83.38} _{\pm 0.13}$\\
& MOON & $74.83 _{\pm 0.68}$ & $85.01 _{\pm 0.09}$ & $66.12 _{\pm 0.71}$ & $84.03 _{\pm 0.28}$ & $53.66 _{\pm 0.14}$ & $82.82 _{\pm0.16}$ \\
&  \textbf{Imba-ULRc }  & $\textbf{87.53} _{\pm 0.40}$ & $85.37 _{\pm 0.06}$ & $\textbf{87.11} _{\pm 0.08}$& $\textbf{85.56} _{\pm 0.13}$ & $\textbf{85.26} _{\pm 0.79}$ & $83.28 _{\pm 0.33}$\\
    \bottomrule
    \end{tabular}}
    
    \label{tab:comperision}

\end{table*}
\par We compare the proposed method with the baseline methods to demonstrate its effectiveness on the three image datasets under different levels of skewness. To ensure a fair comparison between Imba-ULRc and a baseline method, all recovery methods are trained on the same unlearning model when under the same level of skewness. This means that the federated learning process for unlearning is identical, with only the recovery methods differing, to avoid any potential impact on the results from variations in the unlearning model. Each method undergoes global training rounds $\tau=10$, with learning rate and batch size parameters kept consistent with our method. We train the four baseline methods and the proposed method three times each, calculating the average accuracy and standard deviation. The results are shown in Table~\ref{tab:comperision}. As can be seen from Table~\ref{tab:comperision}, the proposed Imba-ULRc method consistently outperforms the four baseline methods in most cases. 
\par For FUDP, the remaining clients directly perform recovery training on the unlearning model without any additional operations or considerations for addressing the label skew issue. It can only leverage the limited information from the remaining clients, where skewed data is easily overshadowed by the other classes. The scarcity of data in the skewed class limits its contribution to the model. Compared to FUDP, the proposed Imba-ULRc method increases the quantity of skewed class data and enhances its quality as well, contributing richer and more comprehensive information from the remaining clients.

\par QUICKDROP balances data by compressing the data from all classes to the same quantity. Although it does not increase the amount of skewed class information from the remaining clients, it maintains balance in the data, preventing the recovery model from completely neglecting the skewed class. However, data compression results in a loss of overall data information and often worsens performance. The compression method also prevents the model from fully learning the data distribution during training, which reduces the model's generalization ability and leads to greater variability in the results. Compared with QUICKDROP, our method preserves the overall data integrity without sacrificing any information.

\par FLRS outperforms the other three baseline methods in the comparison. Although it improves model stability by limiting incorrect updates, it also weakens the model’s ability to capture features of the skewed class, thereby restricting the learning of knowledge specific to this class. This limitation on updates can improve the accuracy of other classes but may allow updates for other classes to overshadow the contributions of the skewed class. However, the proposed Imba-ULRc method not only prevents the skewed class from being overshadowed by updates from other classes but also increases data diversity in the skewed class using SMOTE. This overcomes FLRS's limitations when learning from the skewed class. Besides, Imba-ULRc avoids the issue of restricted knowledge learning for the skewed class, ensuring balanced learning across all classes and further enhancing the overall model performance.

\par MOON aims to reduce the divergence between local updates and the global model, mitigating local training drift caused by skewed distributions to some extent. However, in our scenario, the remaining clients have insufficient data for the skewed class, leading to poor performance of the global model on this class. As a result, the method of constraining local model updates in a specific direction might not be effective here. It could even slow down model convergence, making MOON less competitive than other methods like FUDP, which do not impose such constraints. In contrast to MOON, Imba-ULRc does not restrict the direction of local updates, accelerates model convergence, and reduces the risk of insufficient information in the remaining clients, thereby preventing any hindrance to improvements in skewed class performance.

\par By further analyzing the standard deviation across all methods in the experiments, it is found that although our method introduces some randomness through techniques such as SMOTE, it consistently produces stable results compared to other methods, without causing significant standard deviation. Therefore, the proposed Imba-ULRc method is not only effective in achieving recovery for both skewed and overall classes, but it also exhibits better stability. Besides, it is found that both the accuracies of the existing recovering methods on the skewed class and the global model generally decrease as $\alpha$ increases. For Imba-ULRc, for each dataset, its performances on both the skewed class and the global model remain roughly similar as $\alpha$ increases and are consistently better than the other four methods in almost all cases. This indicates that the proposed method is effective under different degrees of skewness.

\par In summary, the proposed Imba-ULRc method generates data for the skewed class of remaining clients, mitigating the performance bias in the unlearning model. It improves the accuracy of both the skewed class and the overall global model. Furthermore, our method enhances model performance further after denoising. When compared with two existing methods for recovering the unlearning model and two federated learning approaches tailored for skewed label distributions, our method consistently outperforms them in most scenarios.
\begin{table*}

    \captionsetup{format=plain, textformat=period, labelsep=newline, justification=centering}
    \caption{The recovery accuracy of the unlearning model in ablation studies on the proposed method}
    \centering
\scalebox{0.8}{
    \begin{tabular}{ccccccccc}
    \toprule
    Dataset & Method & \multicolumn{2}{c}{$\alpha=0.8$} & \multicolumn{2}{c}{$\alpha=0.85$} & \multicolumn{2}{c}{$\alpha=0.9$} \\ 
    \cmidrule(r){3-4} \cmidrule(r){5-6} \cmidrule(r){7-8}
    & & Skewed class  & Global model  & Skewed class  & Global model  & Skewed class  & Global model  \\ 
    \midrule
    MNIST & Unlearning & 0.00 & 73.42 & 0.10 & 76.75 & 0.10 & 79.81 \\ 
    & RT+SMOTE & 92.80 & 95.66 & 92.41 & 95.86 & 91.54 & 95.86 \\ 
    & RT+SMOTE+DENOISE & \textbf{93.00} & \textbf{95.80} & \textbf{92.70} & \textbf{95.72} & \textbf{91.63} & \textbf{95.88} \\ 
     \midrule
    FMNIST & Unlearning & 0.00 & 50.57 & 0.00 & 54.41 & 0 & 44.18 \\ 
    & RT+SMOTE & \textbf{91.80} &\textbf{ 86.04 }& 87.20 & 85.94 & 88.50 & 85.85 \\ 
    & RT+SMOTE+DENOISE & 89.70 & 85.92 & \textbf{90.00} & \textbf{86.07} & \textbf{88.80} & \textbf{86.05} \\ 
     \midrule
    USPS & Unlearning & 0.00 & 11.31 & 0.00 & 17.84 & 0.00 & 13.40 \\ 
    & RT+SMOTE & 86.39 & 85.33 & 83.67 & 84.65 & 85.03 & 82.37 \\ 
    & RT+SMOTE+DENOISE & \textbf{87.07} & \textbf{85.36} & \textbf{87.07} & \textbf{85.66} & \textbf{85.71} & \textbf{82.90} \\ 
    \bottomrule
    \end{tabular}}
    
    \label{tab:ablation}

\end{table*}
\subsection{Ablation Studies on our Method}

\par To assess whether the proposed method addresses the performance bias and noise removal on the recovery accuracy of the unlearning model, we conducted ablation experiments on three different datasets with three levels of skewness. We report the model accuracy after the unlearning clients executed the unlearning algorithm (i.e., UPGA), as indicated in the ``Unlearning" row in Table~\ref{tab:ablation}. The UPGA method can control the degree of unlearning by adjusting the distance between the model after gradient ascent and the original model. Consequently, our model achieves more comprehensive unlearning on the skewed class. This effectively removes the contribution of skewed class data from the unlearning client, allowing for a clearer and more accurate evaluation of our method's impact on overall performance.

\par 
We validate the effectiveness of the component for rebalancing data (denoted as RT+SMOTE). Each remaining client trains the encoder and decoder for 200 epochs. For those clients, SMOTE is used to over-sample data for the skewed class. We then directly apply the recovery training method introduced in Section 3.4. The results for each dataset are shown in the ``RT+SMOTE" row in Table~\ref{tab:ablation}. It can be seen that using RT+SMOTE only effectively restores the accuracy of the skewed class under different levels of skewness. The accuracies of the global model after recovery indicate that the RT+SMOTE method also effectively improves the performance of the unlearning model on the entire dataset.

    \par Next, we incorporate the data quality enhancement method (described in Section 3.3) into the client-generated data for the skewed class (denoted as RT+SMOTE+DENOISE), to assess their capability to enhance the performance of the recovery model. In our denoising method, we identify $k=5$ neighbors for each data point to compute the density ratio. The results across various datasets are displayed in the ``RT+SMOTE+DENOISE" row in Table~\ref{tab:ablation}. These results demonstrate that recovery training after denoising can enhance the model's performance on the skewed class and a global scale, outperforming RT+SMOTE in 16 out of the 18 cases. The global model’s accuracy also shows a noticeable increase, indicating that RT+SMOTE+DENOISE does not compromise overall quality. This further confirms the effectiveness of our proposed quality enhancement approach.
\subsection{Parameter Sensitivity Analysis}
\begin{figure*}[htbp]
  \centering
  \begin{subfigure}{.33\textwidth}
    \centering
    \includegraphics[width=\linewidth]{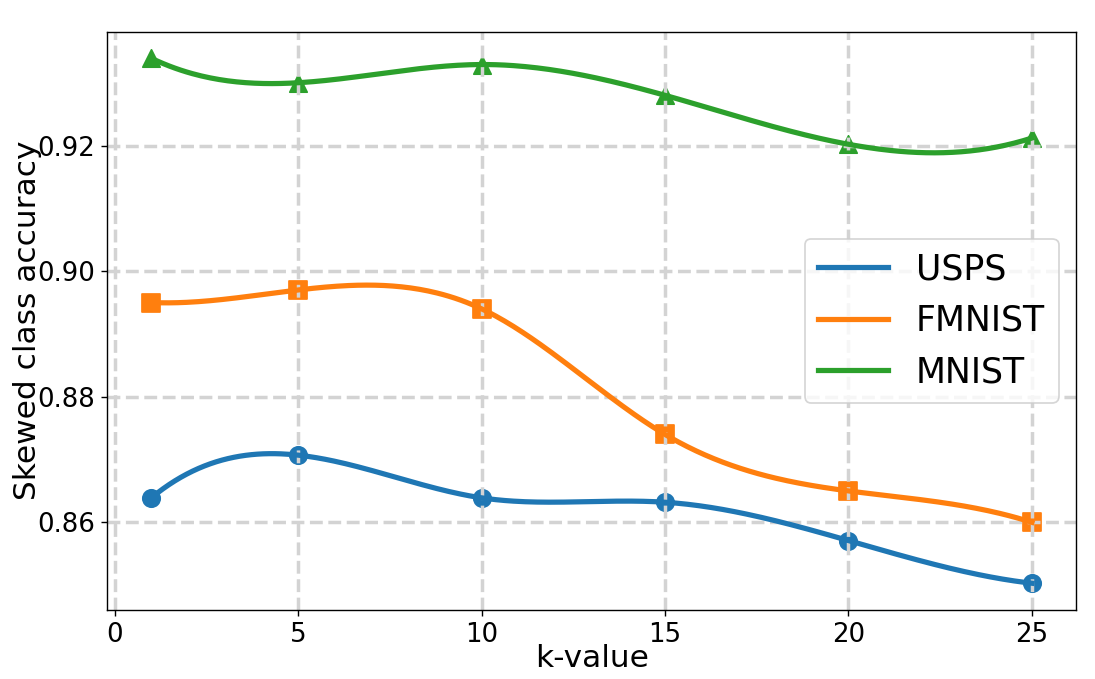}
    \caption{Skewed class accuracy when $\alpha = 0.8$.}
  \end{subfigure}%
  \hfill
  \begin{subfigure}{.33\textwidth}
    \centering
    \includegraphics[width=\linewidth]{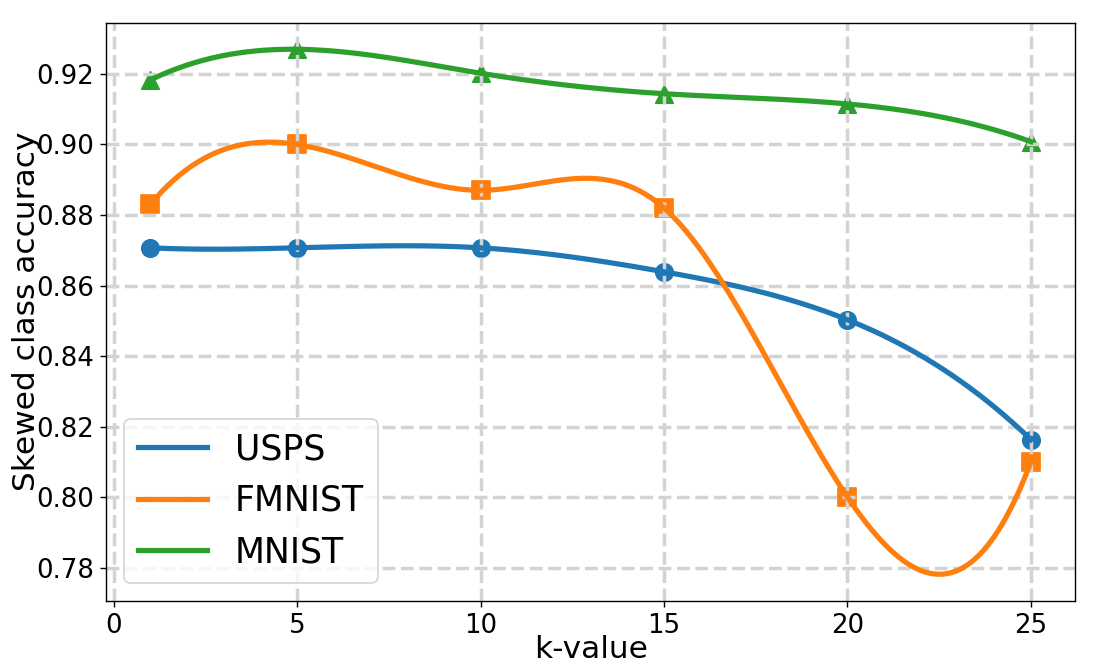}
    \caption{ Skewed class accuracy when  $\alpha = 0.85$.}
  \end{subfigure}
  \hfill
  \begin{subfigure}{.32\textwidth}
    \centering
    \includegraphics[width=\linewidth]{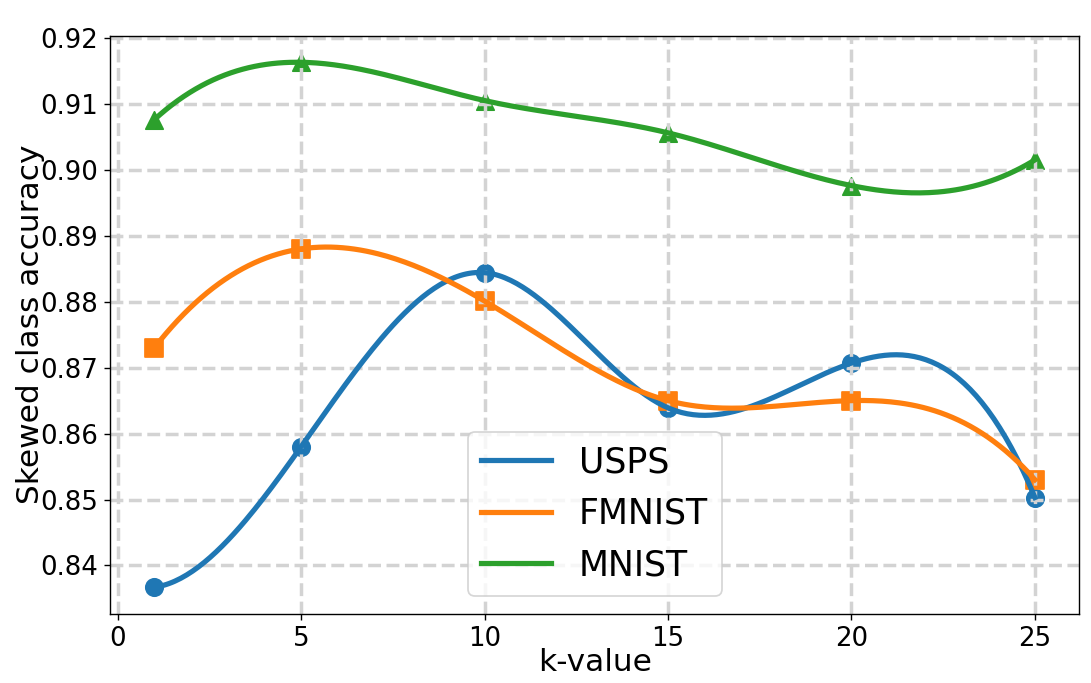}
    \caption{Skewed class accuracy when  $\alpha = 0.9$.}
  \end{subfigure}%
  \hfill
  
 \caption{Under three data skew settings, the accuracy of the model on skewed class after client-generated data denoising and recovery training across various $k$ values on different datasets.}
   \label{K}
  \end{figure*}
 
\par In the noise removal experiment, $k$ is the only parameter that requires manual tuning. The selection of the $k$ value influences the removal of noise in the generated data and the quality of the remaining client datasets. Consequently, we investigate the impact of various $k$ values on denoising effectiveness and identify the optimal setting.
   
\par  On the three datasets in the experiments, we began our analysis with 
$k$ values starting from 1 and increasing up to 25, selecting six distinct values within this range and documenting the accuracy of the skewed class. Figure~\ref{K} presents the balanced accuracy of the global recovery model on the skewed class across the three datasets under various $k$ values and different levels of skewness. It is clear that the data quality enhancement method performs well with lower $k$ values, particularly in the range of [1,10], and almost achieves optimal denoising efficiency at $k=5$. However, as $k$ increases, there is a significant decrease in denoising performance. This is mainly because larger $k$ values, which involve distances to farther points, may decrease the density of normal samples, thereby increasing their likelihood of being incorrectly identified as noise for removal. %This results in the erroneous removal of normal points during the denoising process. 

The graph also shows that the accuracy of unlearning model after denoising data can decrease with the $k$ value increases, even lower than that of only addressing skewed label distribution. However, when the $k$ value is too small, the number of neighbors involved in calculating the local density for each data point becomes inadequate, resulting in an inaccurate estimation of local density. This also causes the estimation to be overly sensitive to individual neighbors, making it susceptible to fluctuations. Therefore, it is reasonable to choose $k=5$ as the optimal number of neighbors for denoising.

\section{Conclusions}

This study aims to improve the performance of the unlearning model in scenarios with skewed label distribution. We achieve it by proposing an unlearning model recovery method. Specifically, when the remaining clients lack sufficient data from the skewed class to effectively recover the unlearning model, we use SMOTE to generate data of the skewed class for these clients, thereby ensuring evenly distributed data across clients in their local datasets. Moreover, since SMOTE is likely to introduce noise, a density-based method is designed to identify and remove noise, further improving the quality of the local datasets of the remaining clients. Finally, all remaining clients use the enhanced local data to iteratively train and recover the performance of the unlearning model. 
\par Experiments on the three datasets demonstrate the effectiveness of our proposed model recovery method. According to the results, by using SMOTE to generate data for the skewed class for the remaining clients, the quality of the unlearning model is improved to some extent. After removing noise, the accuracies of the unlearning model are further enhanced. Compared with the four baseline methods, our proposed method achieves better accuracy for both the skewed class and global models across different levels of data skewness. Although our method is effective in restoring the performance of the unlearning model, further investigations are needed to simplify the recovery process and reduce computational resource overhead. Besides, the recovery process increases the computational burden on the remaining clients, making it desirable to design effective incentive mechanisms to ensure fairness throughout the unlearning and recovery process.

\section*{Acknowledgment}

This work was supported in part by the National Key Research and Development Program of China under grant 2021ZD0112400, the National Natural Science Foundation of China under grant 62206041, Natural Science
Foundation of Liaoning Province under grant 2023-BSBA-030, the Fundamental Research Funds for the Central Universities under grants DUT24LAB122 and DUT24BS010, the Open Fund of National Engineering Laboratory for
Big Data System Computing Technology under grant SZU-BDSC-OF2024-09, the China University Industry-University-Research Innovation Fund under
grant 2022IT174, the 111 Project under grant D23006, 
the Marsden Fund of New Zealand Government under Contract VUW1913, the MBIE Data Science SSIF Fund under the contract RTVU1914, and the MBIE Endeavor Research Programme under contracts C11X2001 and UOCX2104.

\end{document}